\documentclass[12pt]{iopart}

\begin{document}
\title{Constructed emotions and superinformation: a constructor-theoretic approach}
\author{Riccardo Franco}
\address{riccardofrancoq@gmail.com}
\date{\today}
\begin{abstract}
In this paper we apply the constructor-theoretic approach to the theory of constructed emotions, showing that core affect valence and knowledge can be considered as two different observables, leading to information or superinformation conditions: this depends on subject's strategy, coherently with the affect infusion model.
In the second part of the article we show that additional hypotheses on the structure of information allows to study emotions in terms of the contructor-theoretic version of phase task. Quantum algorithms are presented as an example of the connection between emotions and memory tasks.
\end{abstract}
 \maketitle
\section{Introduction}
We know that emotions not only influence the quality of our life, but they also determine the substance of our perceptions and actions, or, in other words, the \textit{context}.  Moreover, they are real, and thus they must correspond to some physical property.  

\textit{Constructor theory} is a new mode of explanation in fundamental physics, first sketched out by David Deutsch \cite{deutsch2013constructor}, expressing physical laws exclusively in terms of what physical transformations are possible versus which are impossible, and why. By allowing such counterfactual statements into fundamental physics, it allows new physical laws to be expressed, for instance those of the constructor theory of information.  The fundamental elements of the theory are \textit{tasks}, abstract specifications of transformations in terms of input/output pairs of attributes. A task is impossible if there is a law of physics that forbids its being performed with arbitrarily high accuracy, and possible otherwise. When it is possible, there exsists a \textit{constructor},  an entity which can cause the task to occur while retaining the ability to cause it again. Examples of constructors include a heat engine (a thermodynamic constructor), a catalyst (a chemical constructor) or a computer program controlling an automated factory (an example of a programmable constructor).

It is clear that such innovative and general approach may find applications also in the study of human cognition and emotions. As written by Lisa Barret, one of the most influential researches in affective science \cite{barrett2007mice}, \textit{every human thought, feeling, and behavior must	be causally reduced to the firing of neurons in the human brain}, or, in more general terms, to specific transitions of physical subsystems in  human brain. Thus, in costructor theory of cognition (which finds a first approach in \cite{franco2019first}), \textit{subjects are constructors of their cognitive tasks},  performing transformations on the brain (considered as a general physical system storing information). This approach results to be very general and it evidences that human judgements make use of a specific type of information, the superinformation. It is clear that such fact may have consequences in modeling human reasoning and artificial intelligence.
The fact that quantum information is a kind of superinformation somehow evidences that the previous attempt to develop a quantum-like theory for human cognition (see for example \cite{busemeyer2011quantum, franco2016newtheory}) was a possible way, even if not fully justified:  there may be other superinformation theories which are required by human cognition.

The use of concepts taken from superinformation in cognitive psychology is new and it may open new perspectives. For example, the study of emotions in cognitive psychology introduces problems and difficulties that superinformation theories may help to treat.  It is clear that our considerations are somehow driven by analogue concepts from quantum computation: even if quantum theory is not required, it may help to guide our attempts to work with superinformation.

\section{Emotions}\label{emotions}
What is an emotion? This question is central to psychology and it is pervasive in scientific models of the mind and behavior. Emotions are seen as the causes, mediators, or effects of other psychological processes such as attention, memory, and perception. Even if  there are unresolved
disagreements over the fundamental question of how an emotion is to be defined, there is general consensus about the definition of emotion in terms of responses to events that are important to the individual: in other words, they are subjective experiences based on feeling pleasure or pain. 
\\
In the following, we briefly present the most important theories and experimental results relevant to emotions, evidencing how the constructor theory of cognition can be used.

Until the 60's, the scientific approaches to emotions focused on the interplay between  body's physiological reactions and emotions, passing from a causal relation (James-Lange theory \cite{james1894discussion, lange1922emotions}  at the end of the 19th-century), to a substantial independence (Cannon-Bard theory \cite{cannon1927james}), up to Schachter and Singer's two factor theory \cite{schachter1962cognitive}.

In the 1960s, Arnold \cite{arnold1960emotion}  developed a new \textit{cognitive approach}, where the first step in emotion is an \textit{appraisal} of the situation (	 the cognitive processes preceding the elicitation of emotion). According to Arnold, the initial appraisal starts the emotional sequence and arouse both the appropriate actions and the emotional experience itself. Thus the physiological changes, recognized as important, accompany, but do not initiate, the actions and experiences. Lazarus \cite{emotlazarus1991emotion}  followed Arnold's approach,  developing a more articulated appraisal theory, where  appraisal mediates between the stimulus and the emotional response, and it is immediate and often unconscious. In contrast to the Schachter–Singer \cite{schachter1962cognitive}  theory of emotions, which views emotion as an outcome of the interaction between physiological arousal and cognition, Lazarus argued that the appraisal precedes cognitive labeling, simultaneously stimulating both the physiological arousal and the emotional experience itself.

Several influential modern models (for example \cite{ekman1999basic}) have preserved the original darwinian assumption that there are different kinds of emotion, each causing a distinct pattern of physiological and behavioral response. In the starkest form of such models, each kind of emotion is biologically basic, a separate, inherited, complex reflex that is hardwired at birth. Many models assume that each emotion kind is characterized by a distinctive syndrome of hormonal, muscular, and autonomic responses that are coordinated in time and correlated in intensity.

However, in these last years, research has progressed from thinking about psychological phenomena as unitary faculties (entities) of the mind to thinking about them as emergent phenomena that vary with the immediate context.  In fact, there is accumulating empirical evidence that is inconsistent with
the view that there are kinds of emotion with boundaries that are carved in nature  \cite{barrett2007mice}. This leads to what Barrett calls the \textit{emotion paradox}, which  evidences that 
our discrete commonsense categories of emotion, like "anger", "sadness", and "happiness" can't find consistent experimental support for their existence.  Instead, the empirical evidence suggests that what exists in the brain and body is a simpler and more primitive element, the \textit{affect}, and emotions are constructed by multiple brain networks working in tandem \cite{barrett2006emotions}. The \textit{theory of constructed emotion} \cite{barrett2006solving} was proposed by  Barrett to resolve such paradox.  She draws an analogy between the categorisation process in emotion perception and colour perception: even if the retina registers light of different wave lengths and the spectrum of wavelengths as a continuum, yet people perceive categories of colours (red, green, yellow, blue) depending on previously acquired conceptual knowledge. The same happens with emotion: whether people categorise an episode of core affect as anger, fear, or sadness depends on acquired conceptual knowledge (emotion scripts).

It is clear that the approach of  \textit{theory of constructed emotion} may have some points in common with a contructor-theoretic approach. In the following subsections we focus on the most important concepts of such theory of constructed emotion. An important question, considered in \cite{clore2007emotions}, is \textit{how emotions inform judgment and regulate thought}.

\subsection{Core affect}
Psychological construction theories hypothesize that emotions are created as the interpretation of affective changes. They integrate dimensional and categorical perspectives, in the following way: it is hypothesized
that all emotional events, at their core, can be described as having psychologically primitive affective properties, which form the “dimensional” aspect of the theories. Psychological construction theories also propose, however, that people automatically and effortlessly use some type of mechanism to these affective changes meaningful in relation to objects and events in the world; this is the “categorical” aspect of
the theories.

According to such view, scientists are able to assess a person’s affective state (i.e., pleasure and displeasure) by more indirect or objective means \cite{cacioppo2000psychophysiology} ,  but these measurements cannot be used to assess feelings of anger, sadness, fear, per se. 
Experimental measurements coordinate around positive versus negative affect \cite{emotcasper2001affective} or intensity of affect, rather than discrete emotion categories. This supports the emerging notion that emotions consist of affective, valenced (i.e., positive and/or negative) reactions to meaningful stimuli, as explicited in theory of constructed emotions \cite{barrett2006emotions}

We can thus move our focus from the list of emotions to the more primitive concept of \textit{core affect}. According to Russell \cite{russell2003core}, core affect is the  \textit{neurophysiological state consciously accessible as the simplest raw (nonreflective) feelings evident in moods and emotions. Core affect is primitive, universal , and simple (irreducible on the mental plane). It can exist without being labeled, interpreted, or attributed to any cause}. As an analogy, we can consider felt body temperature: just like for core affect, you can note it whenever you want; felt temperature exists prior to the concept of temperature, either in folk or scientific theory, and prior to any attribution about what is making you hot or cold.

At a given moment, the conscious experience (the raw feeling) is a single integral blend of two dimensions: 
\begin{itemize}
\item the \textit{ valence}, which summarizes, at the level of subjective experience, how well one is doing,	ranging from a \textit{positive state}   (\textit{pleasure})   to a \textit{negative state}  (\textit{displeasure});
\item  the \textit{arousal} (activation) property, ranging from \textit{maximal activation (agitating)}  to \textit{deactivation (quiescent)}.
At the level of subjective experience, it refers to a
sense of mobilization or energy. A person senses being somewhere
on a continuum ranging from, at the low end, sleep through
drowsiness, relaxation, alertness, activation, hyperactivation, and
finally, at the opposite end. frenetic excitement. Subjective feelings
of activation arc not illusions, but a summary of one's
physiological state. Still, the relation between the relatively simple
subjective experience of activation and its actual complex neurophysiological
substrate is poorly understood; we return to this
topic shortly. Activation
\end{itemize}

The real bipolar structure of core affect has been challeged (see \cite{russell1999core} for a review). Even pleasant and
unpleasant affect, which represent bipolar opposites, may show an empirical independence. We will see in section (\ref{constructor_emotions}) how construcotr theory is able to treat this fact, by using the concept of attributes and variables.

Core affect can be experienced in relation to no known stimulus, in a free-floating form, as seen in moods. However, core affect can be induced by a stimulus. The \textit{affective quality} \cite{russell2003core} is the capacity of such stimulus to change core affect. Perception of affective quality is a perceptual process that estimates this property. It begins with a specific stimulus and remains tied to that stimulus. 

\textit{Attribution of affect} is the perception of causal links between events and allows room for individual and cultural differences. Attributions usually seem correct to the attributor, but research has demonstrated misattributions. Attributed affect is thus defined by three necessary and, when together, sufficient features: (a) a change in core affect, (b) an Object, and (c) attribution of the core affect to the Object.

\subsection{Measurements of emotions}
Even if, from an intuitive  perspective, it should be easy to determine whether someone is experiencing a particular emotion, \textit{scientific evidence suggests that measuring a person’s emotional state is one of the most vexing problems in affective science} \cite{blanchette2010influence}.

We can consider two main types of measurements of emotion.
\begin{itemize}
		\item  \textit{Physiological measurements} (like EEG,  neuroimaging electrodermal and other autonomic studies), which don't  require that subjects conscously evidence their emotions. Actually, it seems that there not exist  autonomic signatures for  discrete emotions: instead, relevant studies often point to relationships among dimensions, particularly those of valence and arousal, and autonomic nervous system responses.	For what concerns EEG, measurements appear to be sensitive to the dimensions of approach and avoidance, but in general neuroimaging methods need more effort to examine interrelated activity among multiple brain regions, since emotional states are complex and involve circuits. 
	
	As noted in \cite{meiselman2016emotion}, variation is  the norm for autonomic measurements taking during emotion.
	Although some writers have made a very persistent case for the existence of autonomic signatures, actual meta-analytic summaries do not support such claims. A meta-analysis \cite{lindquist2012brain}, for example,
	summarized findings from over 200 experiments measuring autonomic reactivity during instances of emotion categories and failed to find distinct autonomic fingerprints for any emotion category.

	\item \textit{Self-reports} of emotion, where subjects have to rate (on a scale of points, eventually also identified by specific labels) their mood (or their satisfaction) after a specific preparation.
	These measurements are likely to be more valid to the extent that they relate to currently experienced emotions. Even in this case, though, there are concerns that not all individuals are aware of and/or capable of reporting on their momentary emotional states. 
	
	We can always describe self-reports measurements  in terms of a judged value $J(A)$ ranging from the minimum to the maximum value of a specific scale, where $A$ is the affect. We can also rescale, if necessary, it in a different range. We can somehow consider such judgements in the same way as other judged values, such as the subjective likelihood of an event. 

In  theory of constructed emotion  there is no “objective”
way of determining when someone is, or is not, in a particular emotional state, and thus the measurement of emotion via self-report is a good technique \cite{barrett2006emotions}. 

Self-reports, on their own, have limitations, of course, because they only capture some instances of emotion (those of which the respondent is aware); in this view, as is the classical view, people may not be conscious of the emotion they have constructed (although for an entirely different set of theoretical hypotheses). But when self-reports do not correlate
with more objective measures, the self-reports are not necessarily assumed to be inaccurate. Moreover, because emotion concepts are integral to the construction of emotional experiences and perceptions, words and other symbols that prime emotion concepts (ie, that launch predictions and simulations) will influence what is experienced and felt. As a consequence, self-reports of emotional experience are influenced by the words that we give respondents to communicate their experiences
or perceptions. It is possible to change a person’s feeling merely by the type of measurement instrument you give them. And respondents will use the measure you give them to report what they want to tell you, which may not necessarily match what you are asking (eg, if a respondent feels excited, but you ask if he/she is happy, the respondent will use the item to tell you how excited he/she is).

\end{itemize}

\subsection{Empirical effects - feeling biases}\label{empirical}
A feeling (or emotional) bias is a distortion in cognition and decision making due to emotional factors. Effects of emotional biases can be similar to those of a cognitive bias, it can somehow be considered as a subcategory of such biases. The specificity is that the cause lies in one's affects, which divert the attention of the person, more than in one's reasoning. Accoding to such  \textit{feeling-as-bias-inducer} paradigm, individuals’ feelings induce various forms of bias into the decision-making process that skew their decisions in certain ways. As summarized in \cite{emotclore2007}, the first robust evidence is the following: 
\\\\
\textit{E1: people's judgments about an event $X$ don't reflect in a clear and unambiguous way information about it: they are  also influenced by information from their own affective reactions. }
\\\\
In other words, any external event $X$ is clearly independent from subject's emotions, but judgements about $X$ are influenced by emotions. 
More specifically experimental manipulations of affect valence (by focusing on specific information or emotions, by designing social situations, or with music) influence in a \textit{mood congruent} way subject's judgements, for example of undesirable events \cite{johnson1983affect, kadous2001improving}, of life satisfaction \cite{schwarz1983mood} or of advertisements \cite{gorn2001arousal}. Such infuence is\textit{ independent of the similarity} between the object $X$ and the element which elicits the emotion \cite{johnson1983affect}. In studies about anxiety, the correlation with judgements seems to be restricted to objects that are self-relevant \cite{muris2006anxiety}: the anxiety-congruent judgements do not extend to events happening to  others.

A second important evidence is relevant to the situations where such affective influence disappears:
\\\\
\textit{E2: the affective influence of judgements is eliminated when subjects focus their attention to their current feelings}
\\\\
This experimental fact evidences that there are specific situations where subjects are able to manage separately information about emotions and about external events. For example, this happens when people attribute their current feelings to the correct causes \cite{forgas2002managing},  if participants are explicitly aware of the source of their feelings \cite{schwarz1983mood} or they are asked to base their evaluations on facts rather than feelings or to rate their anxiety \cite{kadous2001improving}, or when  the judged object has an unambiguous affective tone \cite{gorn2001arousal}.

For what concerns \textit{arousal}, experimental manipulations, independently from valence, lead to more polarized judgements (in direction of the object’s affective tone) \cite{gorn2001arousal}. However, in thisd article we will focus only on valence effects.

\subsection{Affect infusion model}
The Affect infusion model (AIM) is a theoretical model in the field of human psychology, developed by Forgas in the early 1990s \cite{forgas1995mood}.  

Let us consider a generic situat	ion, where subjects have to perform a judgement  about a set of $N$ items: for example, they may have to judge how likely they will found items with a specific property $X$. They have to ckeck each of such $N$ items and count which have this property. If $N$ is big or there aren't strong motivations or resources, subjects may search different strategies. 

The AIM assumes that the nature and extent of mood effects on judgments largely depend on what kind of processing strategy is adopted by a judge. Such strategiies can be classified into:
\begin{itemize}
	\item \textit{low affect infusion} (direct access or motivated processing): they involve direct retrieval of a preexisting stored evaluation or highly selective, guided, and targeted information search;
	
	\item \textit{high affect infusion}  (heuristic or substantive processing): when judges have neither a prior evaluation nor a
	strong motivational goal to help to determine the outcome, they consider only some of the available information, using whatever shortcuts or simplifications are readily available to them. 
\end{itemize}

\section{Constructor theory of information: basic definitions}
Why constructor theory? By focusing on what physical transformations are possible versus which are impossible, constructor theory introduces very useful counterfactual statements into fundamental physics.

Actually, constructor theory is based on the following definitions:
\begin{itemize}
	\item \textit{substrate }($\mathbf{S}$): a physical system some of whose properties can be changed by a physical transformation;
	\item \textit{constructor}: an agent causing the transformation of a substrate, whose defining characteristic is that it remains unchanged in its ability to cause the transformation again;
	\item \textit{state} ($\psi$): a possible configuration of properties of a substrate;
	\item \textit{attribute} ($x$): a set of all the states in which a specific property is true;
	\item \textit{variable} ($X$): a set of disjoint attributes of the same substrate: the specific subsidiary theory will determine how to group attributes into a variable;
	\item \textit{sharp state}: when a substrate is in a state with attribute $x \in X$, we can say that such substrate is sharp in $X$;
	\item \textit{task} ($\mathcal{T}$): a specifications of physical transformations affecting substrates. Different types of transformations will be identified as specific tasks.
\end{itemize}
We recall the basic principle of constructor theory: \textit{all other laws of physics are expressible entirely in terms of statements about which physical transformations are possible and which are impossible, and why}.
Consequently, the element discriminating between two theories will be understanding which tasks are possible and which are not. 

Clearly, a key element in constuctor theory is the task. A constructor is capable of performing a task $\mathcal{T}$ if, whenever presented with substrates having an attribute in $In(\mathcal{T})$, it delivers them with one of the corresponding attributes from $Out(\mathcal{T})$ (regardless of what it does if the substrate is in any other state). On the contrary, the task $\mathcal{T}$ is impossible if there is a law of physics that forbids it  being carried out with arbitrary accuracy and reliability by a constructor.

It is clear that the analysis of possible and impossible task is of fundamental importance in developing subsidiary theories. We can design different specific tasks, but some fundamental are common in all theories:  
\begin{itemize}
	\item \textit{permutation task}: a task which is able to perform, with or without side-effects, a permutation over some set of at least two possible attributes of some substrate:
	\begin{equation}\label{reversible}
	C_{\Pi}(S)\cup_{x\in S}\{x \rightarrow \Pi(x) \};
	\end{equation}
	In case of a binary variable, the permutation task is equivalent to the \textit{NOT} gate or \textit{swap}, which changes 0 in 1 and 1 in 0.
	\item \textit{cloning task}: a task which is able to take  $\{x\}$ and $\{x_0\}$ in  $\textbf{S} \bigoplus \textbf{S}$ , where $x_0$ is  some fixed (independent of $x$) attribute with which it is possible to prepare $\textbf{S}$ from generic, naturally occurring resources, and transform them in two sets  $\{x\}$. In formula:
	\begin{equation}\label{cloning_attributes}
	R_S(x_0)=\cup_{x\in S}\{(x,x_0) \rightarrow (x,x)\};
	\end{equation}
	\item \textit{distinguishing task}: a task which maps a variable into another clonable one: 
	\begin{equation}\label{task_distinguishable}
	D(S)=\cup_{x\in S}\{x \rightarrow i_x\};
	\end{equation}
	Being able to distinguish between different attributes is important. Moreover, it allows to introduce the \textit{bar}-operation $\bar{x}=\bigcup_{a: a \perp x}a$ which is the union of all the attributes distinghuishable from $x$. Another definition based on the notion of distinguisghability is the attribute $u_{X}=\bigcup_{x \in X} x$. Following the example of the Ellsberg paradox \cite{segal1987ellsberg}, let us consider an urn containing 30 black balls and 30 yellow balls: $u_X$ describes the state when we extract a ball from the urn, since it is the union of each specific possible extractions. Moreover, $\bar{\bar{u}}_{X}$ represents the generalization in constructor theory of the notion of subspace. Following the same example, we now don't how many black or how many yellow balls there are, but the total number of black balls plus the total number of yellow equals 60.  In this case, $\bar{\bar{u}}_{X}$ represents not only all the possible configurations of yellow and black balls, but also, when admitted in the model, other emotional states, consistently with the concept of knittian uncertainty (see for example  \cite{sarin1998revealed})	.
	
	\item \textit{measuring task}: this task describes in constructor-theoretic terms the measurement operation.
	\begin{equation}\label{task_measurable}
	M_S(x_0)=\cup_{x \in X} \{(x,x_0)\rightarrow (y_x, 'x') \};
	\end{equation}
	When $X$ is sharp, the output substrate ends up with an information attribute $x$ of an output variable, which
	represents the abstract outcome ‘it was $x$’.  On the contrary, if the initial state is non-sharp, the result of the measurement can be one of the possible values of $X$.
	\item \textit{preparation task}:  given a substrate $\mathbf{S}$ with a binary variable $X$ with attributes $x_0,x_1$, the \textit{preparation} or \textit{fixed} task is
	\begin{equation}\label{Timp1}
	\mathcal{T}_{x}=\cup_{\psi \in S}\{(x_0,\psi) \rightarrow (\theta_x, x) \}
	\end{equation}
	whose meaning is simple: the initial state $\psi$ changes into $x$, while $x_0$ is a generic, ‘receptive’ attribute in another substrate which interacts with the other becoming $\theta_x$ (after the preparation, such substrate can be discarded). 
	\item \textit{identity task}: it leaves unchanged any state and each attribute $x\in X$ for every variable $X$ of the same substrate. It can be applied both on superinformation and on classic information media.
\end{itemize}
Tasks may be composed into networks to form other tasks, as follows. The parallel composition $\mathcal{A}\otimes \mathcal{B}$ of two tasks $\mathcal{A}$ and $\mathcal{B}$ is the task whose net effect on a composite system is that of performing $\mathcal{A}$ on $M$ and $\mathcal{B}$  on $N$. When $Out(\mathcal{A})=In(\mathcal{B})$, the serial composition $\mathcal{BA}$ is the task whose net effect is that of performing $\mathcal{A}$ and then $\mathcal{B}$ on the same substrate. A regular network of tasks is a network without loops whose nodes are tasks and whose lines are their substrates, where the legitimate input states at the end of each line are the legitimate output states at its beginning. Loops are excluded, because a substrate on a loop is a constructor.

In constructor theory we can identify specific situations based on the allowed tasks. This helps us to quickly specify the actual situation.
\begin{itemize}
	\item \textit{computation variable}: a set of two or more attributes
	for which a reversible computation is possible;
	\item \textit{computation medium}:  a substrate with at least one computation variable;
	\item \textit{information variable}: a clonable computation variable;
	\item \textit{information medium}: a substrate that has at least one information variable;
	\item \textit{information observable}: an information variable $X$ for which, whenever a measurer of $X$ delivers a sharp output ‘$x$’, the input substrate really has the attribute $x$.  Observables have the property that $\bar{\bar{x}}=x$. 
	 \item \textit{superinformation}: an information medium with at least two information observables that contain only mutually disjoint attributes and whose union is not an information observable. 
\end{itemize}

\subsection{Non-sharp states}\label{non-sharp}
Clearly working only on sharp states is not very useful. A strategy for studying non-sharp states, described in \cite{marletto2016constructor}, is to consider multiple instances of the observable $X$ of a substrate $\textbf{S}$.  In fact, since an observable is also an information variable, we can consider multiple instances of it. 
Thus we can define a \textit{counting task} $\overrightarrow{D}(N)_x$ which outputs the fraction of instances of $S$ on which $X$ is sharp with value $x$.  If $N \rightarrow \infty$, the attributes output of $\overrightarrow{D}(N)_x$ define the \textit{X-indistinguishabilitiy} class (the set of all attributes with the same \textit{X}-partition of unity): any two attributes within that class cannot be distinguished by measuring only the observable $X$ on each individual substrate, even in the limit of an infinite ensemble.

If all the measurements give us the same value $x$, we have a \textit{X-indistinguishability equivalence class} denoting the sharp state $x$. On the contrary, if, given a state $z$, different measures of the observable $X$ lead to different values, we have a non-sharp state and a particular $X$-indistinguishability class. We call $z$ a generalized mixture of $X$ when it is indistinguishable for any $x\in X$ and $\{\bar{\bar{u}}_{X}, u_{X}\}$ is sharp in $z$.  

These indistinguishability classes allow us to define the concept of $X$\textit{-partition of unity} as $[f_x]_{x\in X}$, with $f_x \in [0,1]$ and $\sum_x f_x =1$ for any state of the substrate, even for generalized mixtures. The real numbers $f_x$ are not probabilities at ther moment, they  only form an array of real numbers whose sum is 1, thus defining labels of equivalence classes. 

\subsection{Superinformation}
We recall from \cite{marletto2016constructor} the definition of \textit{superinformation: an information medium with at least two information observables that contain only mutually disjoint attributes and whose union is not an information observable}. 

Superinformation shows some important properties, as described in \cite{marletto2016constructor}: 1) not all information attributes of a superinformation medium are distinguishable; 2) it is impossible to measure whether the observable $X$ or $Y$ is sharp, even given that one of them is; 3) superinformation cannot be cloned; 4) pairs of observables are not simultaneously preparable or measurable; 5) superinformation media evolve deterministically yet unpredictably; 6) measuring one observable causes an irreducible perturbation of another; 7) for any arbitrary subsidiary theory that conforms to constructor theory, consecutive non-perturbing measurements of an observable, even if it is not sharp, yield the same outcomes.

Clearly, it is very important to identify situations where the substrate supports superinformation or information. In quantum physics, the union of two sets of  orthogonal states of a qubit instantiates superinformation.
In \cite{franco2019first} it is shown an example, taken from a famous cognitive bias, the conjunction fallacy, where human substrates managing data instantiate superinformation. We can see that models using superinformation generally share the same puzzling features.

We now prove a simple general property of information and superinformation: given two   observarvables $X$ and $Y$ of an information medium, we have the following property of the relevant to the partitions of unity:
\begin{equation}\label{conj_general}
	\forall x\in X, y \in Y: f_{x, y} \leq f_x, f_{x, y} \leq f_y  
\end{equation}
This can be demonstrated by noting that for an information medium we can measure simultaneously both $X$ and $Y$, for any copy of the original substrate. Thus if we count the number of copies where specific values $x$ and $Y$ are observed, such number is always lower or equal than the number of copies for only a single attribute $x$ or $y$. Cleary, when condition  \ref{conj_general} is violated, we have a superinformation medium.

\subsection{Conditional task}\label{task_conditional}
A \textit{conditional task} for two substrates $S_1$ and $S_2$ acts for example on a variable $X^{(2)}$ of the second substrate conditionally on specific values of a variable $X^{(1)}$ in the first substrate:
\begin{equation}\label{conditional}
\mathcal{CT}= \{(0,\mathcal{I}(X^{(2)}), (1,\mathcal{T}(X^{(2)})))\}
\end{equation}
where $\mathcal{I}$ is the identity task for the second substrate, while $\mathcal{T}$ is a generic task for the same substrate.
The simplest example is the $\mathcal{CNOT}$ task for binary variables, where  $\mathcal{T}$ is a swap task.

We now consider conditional tasks for classical information  and superinformation media.
\begin{itemize}
\item In classic information case, the conditioning task for two events $X^{(1)},X^{(2)}$ in two different substrates  is a task which acts on the second substrate only if the attribute of $X^{(1)}$ has a specific value, for example $x_1^{(1)}$.
\item In superinformation case, we have in each substrate two complementary observables $X^{(1)}, Y^{(1)}$ and $X^{(2)}, Y^{(2)}$. The conditioning task could be driven in the first substrate from $X^{(1)}$ or $Y^{(1)}$ and similarly in ther conditioned substrate. 
In the following, we consider different conditioning situations: \textit{direct conditioning}, where $X^{(1)}$ conditions $X^{(2)}$ (or similarly  $Y^{(1)}$ conditions $Y^{(2)}$), and \textit{indirect conditioning}, with observables $X^{(i)} Y^{(j)}$ involved (with $i \neq j$). These kind of conditioning are very important in constructor theory of emotions, since observables $X$ and $Y$ will represent the knowledge of an external event and the core affect valence respectively.
\end{itemize}

	\section{Constructor theory of cognition}\label{constr_cognition}
We now briefly recall the main points of constructor theory of cognition, as originally intrdouced in \cite{franco2019first}. 

\begin{itemize}
	\item The first hypothesis is that \textit{every human thought or emotion is represented by neurophysiological states, which are specific physical configurations in the brain}. The \textit{substrate} is the part of human brain where such thoughts are encoded and managed. It is not importsant in this context if such portion of brain is localized or widespread in the brain and we don't reduce cognitive processes simply to the firing of neurons in the human brain. We simply state that thinking or knowing something corresponds to different neurophisiological states.
	\item The \textit{attribute} is a  neurophysiological state with a specific property (in cognitive terms): for example, knowing something, being happy, thinking to something. This very general approach of constructor theory helps us to carefully evidence additive hypotheses of the model.
	\item The \textit{variable} is a  group of attributes designed in a convenient way to manage the knowledge of facts. For example, we identify as $X$ the binary variable describing the knowledge that an event $X$ is true or false. The structure of a variable may depend on the context: for example, 	different colors can be consdered as attributes of the variable \textit{color}. However, it is known that in different cultures colors are categorized in different ways.  We will see that variables are important salso in the study of emotions: in some cases, attributes relevant to the affect valence can be grouped to form a variable, while in other cases they don't form a vartiable.\\
	Generally speaking, it is not convenient to group in the same variable attributes which are difficult to manage together. In the following, we will add definitions which will help in this perspective. 
	\item \textit{Any human change of thought or feeling can be made in correspondence with tasks of constructor theory}. In fact, tasks are transformations of substrates. 
	\item \textit{Every subject is a constructor for his cognitive tasks}.
\end{itemize}

The \textit{basic principle }  of constructor theory, reframed in cognitive context, states that \textit{all the laws of cognitive science are expressible entirely in terms of statements about which physical transformations are possible and which are impossible, and why}. Clearly, the constructor theory helps us to distinguish between different situations where some tasks are possible or not.

The \textit{second principle} \cite{marletto2016constructor} states  in constructor-theoretic form Einstein’s principle of locality:  \textit{there exists a mode of description such that the state of the  combined system  of any two substrates $S_1$ and $S_2$ is the pair $(x, y)$ of the states $x$ of $S_1$ and $y$ of $S_2$, and any construction undergone by $S_1$ and not $S_2$ can change only $x$ and not $y$}. This means in cognitive terms that different subsystems in the same brain or in different subject's brains can be changed without influencing reciprocally. 

We now recall a third principle which has been conjectured in constructor theory of information, the \textit{interoperability principle}, which states that \textit{the combination of two substrates with information variables $X_1$ and $X_2$ is a substrate with information variable $X_1 \times X_2$}. In cognitive terms, this means that subjects can organize the brain subsystems, working with these combinations. For example, the state $\psi$ of a subject can be the collection of different attributes of many variables $\psi=(x_1,x_2,...)$ which describes the fact that in the same time the subject is able to take into account different elements of knowledge, affect...

We now review the previously defined tasks in  cognition context. 
\begin{itemize}
	\item \textit{permutation task} $C_{\Pi}(S)$:  it describes the human capability to change the focus on different concepts in a given set. For example, thinging to color red, then blue and then again red.  A substrate for which permutation is possible on at least one variable is a computation medium;
	\item \textit{cloning task} $R_S(x_0)$: it allows to replicate one neurophysiological state of a portion of brain on other portions. It is clearly of fundamental importance  in the rational reasoning, where subjects can rethink to something and cosider it from a different point of view. When a variable is clonable, we have an \textit{information medium}. On the contrary, when a variable is not clonable, the subject can't replicate such variable, which is fleeting;
	\item \textit{distinguishing task} $D(S)$: when possible, it  remaps the knowledge of something in another frame which allows cloning. Distinguising it is at the basis of metaphores and conceptual mapping. For example, when we have a binary variable, we often describe it in terms of \textit{on/of},\textit{ok/ko} or ${0,1}$, clearly using variables from other contexts. In these cases, we reframe the binary variable in terms of other simple well-known categories. Viceversa, a variable is not distinguishable when the subject cannot manipulate it in a replicable way. It is well known in psychology that being able to reframe requires statring with a sufficiently simple set of attributes;
	
	\item \textit{measuring task} $M_S(x_0)$: it is the process of reading information from one substrate and encoding it into another.  The definition (formula \ref{task_measurable}) is based on its action over the set of attributes $\{x\}$ of variable $X$: the measurement of such states results in encoding in the second substrate the response '\textit{it was x}'. For such a reason, these  states are also called \textit{sharp} in $X$ and they represent situations of clear knoweledge about $X$. We say that a variable $X$ is \textit{measurable} when the measuring task $M_S(x_0)$ of formula (\ref{task_measurable}) exists: when the subject knows that $x$ is true, he is able to answer coherently.
	\\
	When a subject is \textit{uncertain} about an event, we have a \textit{non-sharp state} for the variable corresponding to such event: in this case, repeated measurements of the variable lead to different possible results, which means that many copies are sampled to measure the actual value of the variable. \\
	There are  concepts that are by definition difficult to be measured  univocally, while there are variables such that whenever the measurement delivers a sharp output the input substrate really has that attribute: in this case we have an \textit{observable}.
\end{itemize}

\subsection{Judgements and bounded/full rationality regime}
The concept of $X$-indistinguishability class defined in constructor theory is useful in the corresponding cognition theory because it offers a simple tool to manage non-sharp states. We recall that in cognitive context such non-sharp states represent situations where a subject is uncertain about a specific event. Thus the set of non-sharp states in the same $X$-indistinguishability class define the cognitive states of uncertainty about $X$. Moreover, it is also interesting to consider how such class is build: a cloning task is performed on the original state, and a measurement is made on each copy: in cognitive context, this can be interpreted as a \textit{sampling} over copies of the same state. In other words, we can imagine that the mental representation of an uncertain event is replicated many times and tested. Judgement $J(x)$ about an uncertain event $x$ is performed by repeated measurements over a large number of copies of the original state, and by finally counting the number of copies where the event results to be true. In other words, judged likelihoods are, under precise hypotheses, equivalent to the $X$-partitions of unity $f_x$ defined in section (\ref{non-sharp}).

As evidenced in constructor theory, judged likelihoods $J(x)$, just as  the values of the $X$-partition $f_x$, aren't in general probabilities, but \textit{only labels of equivalence classes}. Additional conditions for them to inform decisions in the way that probabilities are assumed to do in stochastic theories are given via the decision-theory argument \cite{marletto2016constructor} and will be recalled later in this article.

In this section, we simply remark the main point of \cite{franco2019first}: constructor theory, when applied to cognition, helps us to distinguish between a regime of (full) rational reasoning and a bounded rationality regime. In the first case, each observable and combination of observables is an information medium, and we can perform cloning task on each observable, allowing for rational judgements on any uncertain element. In bounded rationality regime, there are couples of observables whose union is not an information observable: in this case, we can say that specific physical subsystems of the brain (that substrate) is a superimfornation medium. In this regime, we can't judge correctly any observable.

\subsection{Superinformation effects in cognition}\label{inf_superinf}
In \cite{franco2019first} it is shown an example, taken from a famous cognitive bias, the conjunction fallacy, where human substrates managing data instantiate superinformation. We can see that models using superinformation generally share the same puzzling features.
We now evidence the most important consequences of managing superinformation in cognition. This will be useful in the next section, when applied to emotions, to prove that emotions require spuerinformation.

We provre now the following \textit{conjunction property}: \textit{let us consider two information observables $X$ and $Y$ that contain only mutually disjoint attributes and whose union $X \cup Y$ is  an information observable. Then, if subjects are able to perform a counting task (and thus produce judgements), we have:}
\begin{equation}\label{conj_general}
\forall x\in X, y \in Y: J(x, y) \leq J(x), J(x, y) \leq J_y  
\end{equation}
This propoerty can be easily explained: let be $\{x_i\}$ and $\{y_j\}$ the attributes of observables $X$ and $Y$: then we can perform a cloning task, whose action for any attribute $x_i,y_j$ is:
\begin{equation}\label{cloning_attributes_2obs}
\{(x_i,_j,x_0,y_0) \rightarrow (x_i,y_j,x_i,y_j)\};
\end{equation}
This means that subjects can manage both observables by manipulating new copies to perform measurements and rational judgements: thus we can measure simultaneously both $X$ and $Y$, for any copy of the original substrate. If we count the number of copies where specific values $x$ and $y$ are observed, such number is always lower or equal than the number of copies for only a single attribute $x$ or $y$. Cleary, \textit{when condition  (\ref{conj_general}) is violated and $X$ and $Y$ are observables, we have a superinformation medium} (supposing that subjects are able to perform counting task).

We recall now some important properties of superinformation medium and an interesting prediction. First of all, pairs of observables $X$ and $Y$ are not simultaneously preparable or measurable: this means that subjects can't know in the same time such elements. Secondly, not all information attributes of a superinformation medium are distinguishable: this means that subjects, when managing two complementary variables $X$ and $Y$, can't distinguish them: they are preceived as confused. This fact leads to an important prediction at neurophysiological level: we expect that neuroimaging techniques can't distinguish different variables when managed in a superinformation medium. On the contrary, the same variables, in a context of classical information medium (thus a rational reasoning), will be distinguishable with the same neuroimagin techniques.

\subsection{Categorization}
Categorization is an activity that consists of putting things (objects, ideas, people) into categories (classes, types, index) based on their similarities or common criteria. It allows humans to organize things, objects, and ideas that exist around them and simplify their understanding of the world.

The catgorization task, given a specific category, can be described in the constructor theory as a modified measurement task
$Cat_c(x,x_0)=\cup_{x \in X} \{(x,x_0)\rightarrow (y_x, \chi_c(x)) \}$, where $\chi_c(x)$ is a boolean function which is true only when  attribute $x$ is true for the category item $c$.  For example, variable $X$ may identify a  feature, and only some specific values of this feature may be true for category $c$.
When the  state $s$ is non-sharp, the categorization task can't give a unique answer: the concept of $X$-indistinguishability classes can be applied to $\chi(X)$, leading to the concept of category judgement $J_c(s)$.

Cleary, the structure (according to the constructor theory) of the variable $X$ on which the categorization process is applied is of fundamental importance. If $X$ is an observable, then $J_c(s)$ can be interpreted as a classic similarity judgement. However, if we consider a collection of variables $X\cup Y$ forming a superinformation medium, then the process of counting the copies for which $f_c$ is true is problematic, leading to situations similar to the conjunction fallacy (for example, the well known pet-fish problem, or guppy effect \cite{smith1984conceptual}).

\subsection{Substrate addition and control of coherence}
As already evidenced, we can have, in the same time, different thoughts: for example we can think to our work and simulaneously keep in mind something other.  Thus the state $\psi$ of a subject can be the collection of different attributes of many variables $\psi=(x_1,x_2,...)$ which represent different elements of knowledge, affect...	

As pointed out in \cite{franco2016newtheory},  when a subject answers to a question, the measurement process encodes in a second substrate the answer, coherently with the constructor-theoretic definition of measurement. This contrasts with earlier quantum-like cognition theories, where the subject's answer simply leads to a collapse of the original state. 

From a psychological point of view, a subject may know a specific situation, think differently and even answer to a question about it in a completely different way. This means that variable $X$ relevant to the knowledge of an event may be accompanied by another variable $X'$ (what the subject hopes), and a third one $X''$ (what the subject says).

Thus the cognitive system becomes larger, and needs of some internal mechasnism of coherence. In fact, after the answer has been encoded in a specific substrate, the subject may receive update information or continue thinking and modify the original state, originating the \textit{cognitive dissonance} \cite{festinger1962theory}. 
The presence of internal coherence mechanism is hypothesized also in emotion studies \cite{forgas2002managing}, in the form of homeostatic cognitive strategies in affect regulation.

\section{Emotions in constructor theory of cognition}\label{constructor_emotions}
We first assume the same definitions and hypotheses already taken is section (\ref{constr_cognition}): in particular, every cognitive or affective state is constructed by a constructor (the subject) on a specific substrate. It is easy to recognize that such constructor theoretic assumption is highly compatible with the approach of theory of constructed emotions \cite{barrett2006emotions, russell2003core}. Recalling Russel's definition \cite{russell2003core},  \textit{core affect}  is the  \textit{neurophysiological state consciously accessible as the simplest raw (nonreflective) feelings evident in moods and emotions. Core affect is primitive, universal , and simple (irreducible on the mental plane.}. At a given moment, the conscious experience (the raw feeling) is a single integral blend of two dimensions: the valence, ranging from pleasure to displeasure, and the activation. 

As evidenced in theory of constructed emotions, affective feelings  are experiential representations of value: an emotion is an affective state that represents appraisals of something as good or bad: Russell calls it \textit{valence}, which may assume values from \textit{displeasure} to \textit{pleasure}, with different granularity. 

We remember that in constructor theory an \textit{attribute} is simply a state denoting a particular configuration of the substrate. Thus we can denote as $'a'$ an attribute describing the neurophysiological state  where the subject has a specific affect valence: for example $a$ can be \textit{pleasant state}. 
Grouping attributes leads to the concept of variable. However, there may be different kinds of groups of attributes, some of them with a menaning, other without a meaning. As evidenced in section (\ref{emotions}), in specific situations \textit{core affect valence} can be descibed in the simplest form as a  \textit{binary} variable $A$ with two sharp states: $a_{+}$ is the positive valence, representing pleasure,  while $a_{-}$ is the negative valence, representing displeasure. 

We assume that valence is also measurable: this means that subjects in a specific valence state $a$ are able to answer $'a'$ (\textit{I'm with affect valence a}). Of course, subjects can also be in non-sharp states: the measurement over different copies may give different values: sometimes \textit{pleasant}, sometimes \textit{unpleasant}. The counting task helps to define the \textit{judged valence} $J(a)$ describing the judged perception of valence. This judgement can assume values on a specific scale (for example, from 0 to 100) or rough judgement in terms of labels (like form example \textit{completely good}).

Similarly, we recall from constructor theory of cognition \cite{franco2019first} that the knoweledge of an event can be defined as a binary observable $X$ relevant to the \textit{knowledge} (or belief) about it  $X=\{x_0,x_1\}$. For example, \textit{the coin is tail}, or \textit{Linda is bankteller}, or \textit{Bob plays the guitar}. They are binary events which can be verified and that can have unambiguous mental representations, according to the constructor theoretic definitions.

\subsection{The appearence of superinformation}\label{affect-superinformation}
Now that we have two observables  $X$ (the object) and $A$ (the affect valence, or simply the affect), we can ask which is the relation between them. If is known that core affect can be \textit{object-free} or \textit{object-oriented}. In the first case, we also call it \textit{mood}. In the second case \textit{knowing}  and \textit{affect} become \textit{tightly linked}, and disentangling them requires methods by which affect can be varied independently of belief \cite{clore2007emotions}. 

Coherently with the conceptual tools introduced by the affect infusion model (AIM), we note that in low affect infusion strategies affect can be managed as an object-free element, even if it can produce indirect influence. On the contrary, in high affect infusion strategies, affect is object-oriented and tightly correlated with judged events. We now recognize from section (\ref{empirical}) that evidence $E1$ can be formalized as:  
\begin{equation}
J(x)\neq J(x|z),
\end{equation}
where variable $X$ is the judged object, $A$ the affect valence, and $z$ a generalized mixture of $A$. In other words,  subject's  judgements about $x$ is modified if it is conditioned by an affective state $z$. 
However, the second evidence $E2$ shows that the explicit knolwedge of affect (that is the sharp state $a$) leads to
\begin{equation}
	J(x)=J(x|a)
\end{equation}
In fact, it is known that the simple act of consciously focusing on emotion leads to evidence E2. We now prove a new relevant fact for information medium, the following \textit{independence property}: given a generalized mixture $z$ of an observable $A$ forming an information medium with $X$:
\begin{equation}\label{cond_general}
\forall x\in X, a \in A:  J(x)=J(x|a) \Rightarrow J(x) = J(x|z)  
\end{equation}
In other words, \textit{for an information medium, if an observable is independent for any sharp conditioning, then such independence must hold also when $A$ is unsharp}.  In brief, for an information medium we have that $E2 \Rightarrow \neg E1$. This fact can be explained by noting that, since $X$ and $A$ form an information medium, they can be both cloned and measured simultaneously. Thus we can consider multiple instances of substrate $S$ for such observables and perform measurements on them, leading to $X$ (and $XA$)-partitions of unit and to judgements $J$. Thus, since $z$ by definition is the union of possible states $a$ and by hypothesis $J(x)=J(x|a)$ the second condition $ J(x)=J(x|z)$ must be true. 

The consequence of property of formula (\ref{cond_general}) is that, when it is violated, we have a superinformation medium. We easily recognize from section (\ref{empirical}) that, in many high affect infusion situations, events  $E1$ and $E2$ are true, thus  entailing that the description of emotion experiments requires the use of superinformation.  In the context of the affect-as-information hypothesis, we can conclude that affect valence is managed together with other events as an information variable in low affect infusion situations, and as superinformation in high affect infusion situations.

\subsection{Conditions for decision-supporting superinformation theories}\label{decision}

Now that we know that in high affect infusion strategies the affect valence $A$ and the judged  event $X$ form a superinformation medium, we look for more specific relations between such observables. 

Let us start with a simple example: when we say \textit{the glass is half full/empty}, we refer to two different emotional approaches to the same situation of uncertainty about something (for example event $X$). We can have a \textit{positive affect} approach, where the glass is half-full, or a \textit{negative affect} approach for the half-empty. From a rational point of view, half-full/empty states are without of meaning: in the hypothesis that $X$ is a binary observable (empty/fulll), we have a unique unsharp state with   $X$-partition of identity  $[f_{x_0},f_{x_1}]=[\frac{1}{2}, \frac{1}{2}]$.
On the contrary, the presence of affect allows us to define  different non-sharp states of observable $X$ in terms of atrributes $a_+$ and $a_-$, the half-full and half-empty states.  

The previous example leads us to require as additional hypotheses the conditions for decision-supporting superinformation theories, formalized in \cite{marletto2016constructor} : they are the additional sufficient conditions  for the partition function $f_x$ to inform decisions in the way that probabilities are assumed to do in stochastic theories (including traditional quantum theory via the Born Rule). In other words, these define under what circumstances the numbers $f_x$ can be used as judged likelihoods $J(x)$ if they were probabilities. Such conditions are given by introducing the concept of generalized mixtures of  two  binary observables $X = \{x_1, x_2\} , A = \{a+, a-\} $ with the following symmetry requirements  used in  the decision-theory argument \cite{marletto2016constructor}:\\
\begin{itemize}
\item \textit{$R1$: $x_1, x_2$ are generalized mixtures of $A$ and $\{a_1, a_2\}$ are generalized mixtures of attributes of $X$}. This means  that  we consider on the same substrate two observables: each attribute of one variable is a generalized mixture of the attributes of the other variable.

This condition describes a specific situation of object-oriented emotion: we can't build affect states (about event $X$) on substrates separated from those relevant to the knowledge of event $X$.

\item $R2$. $S_{x_1,x_2} (a_{\pm} )\subseteq a_{\pm} ; S_{a_+,a_-} (x_i) \subseteq x_i, i=1, 2$, where the computation $S_{m,n}=. \{m \rightarrow n, n \rightarrow m\}$  swaps the attributes $m, n$ of $S$. This means that these mixture are symmetric by swapping thre constituent attributes.

In emotion theory, this condition requires that we can build couples of observables forming superinformation which are symmetric under swap task: for example in the empty/full glass, simply exchanging words \textit{empty} with \textit{full} leads to a swap of the states (from pessimistic to optimistic). 

\item $R3$: the $A$-partition of states $x_0,x_1$ is such that $[f_{a_+}]_X=[f_{a_-}]_X$: this means that  each attribute of one variable is a generalized mixture of the attributes of the other variable  with equal weights: swapping the constituents of such mixtures doesn't change the indistinghuishability class of the mixture.

In the context of constructed emotion, this condition allows to build affect states $a$ with complete ignorance about variable $X$, and similarly knowledge states $x$ with non-sharp affect states.  Moreover, the state describing the knowledge of $X$ represents an unsharp state for affect. Similarly, a sharp affective state is an unsharp knowledge state. 

\item $R4$: conditions $R1, R2, R3$ can be applied also with substrates $S' \oplus S''$ with observables $X=(X', X'')$ and $A=(A', A'')$. There exists an attribute $q$ that is a generalized mixture of attributes in $X$ and a generalized
mixture of attributes in $Y$  such that $S_X(q) \subseteq q$ and $S_A(q) \subseteq q$ .
\end{itemize}

In the following, we assume that these conditions are true, allowing us to consider the $X$-partition of unity as true probabilities and affect as an unsharp state relevant to $X$. In the binary case, we can write the $X$-partition of unity $[f_{x_0},f_{x_1}]$ as a function  of a unique parameter, that is:
\begin{equation}\label{theta}
[f_{x_0},f_{x_1}]=[cos^2(\theta/2), sin^2(\theta/2)].
\end{equation} 
For values of parameter $\theta=2k\pi$ (with $k$ integer), we have the sharp state $x_0$, while for $\theta = (2k+1)\pi$ we have the sharp state $x_1$. Given $\theta \in (0,\pi)$, we have that $\theta$ and $\theta +\pi$ represent two different non-sharp states in the same $X$-indistinguishability class.

Thanks to the decision-supporting hypotheses, we can consider such parameter $\theta$ as a label for indistinguishability classes.  When $\theta= (2k+1)\pi/2$, we recover the two attributes $a_+$ and $a_-$, which are generalized mixtures with equal weights described in condition $R3$.

In conclusion, in section (\ref{affect-superinformation}) we have shown that affect requires superinformation, and in this section we added reasonable hypotheses to describe affect.

\subsection{The coin toss and the W tasks}
Given a boolean observable $X$, we look for a \textit{trasformation which transforms any attribute $\{x_0. x_1\}$ into a non-sharp state in an $X$-indistinguishability class with equal weights for $x_0$ and $x_1$}. In other words, it transforms the sharp states $x_0,x_1$ into general mixtures whose $X$-partition of identity is $[f_{x_0},f_{x_1}]=[\frac{1}{2}, \frac{1}{2}]$, that is a change of parameter $\theta$ of $\pi$.
\begin{itemize}
	\item  In the case of classical information, there is a unique state $\mu$ whose $X$-partition of identity is $[f_{x_0},f_{x_1}]=[\frac{1}{2}, \frac{1}{2}]$. Thus we can call such task the \textit{coin task} $\mathcal{C}(x)$, which acts in the following way: $x_0 \rightarrow \mu$ and $x_1 \rightarrow \mu$. The transpose of $\mathcal{C}(x)$ doen't exist, since it would be a multivalued function.
	\item  In the case of superinformation, there is another observable $A$ such that $X$ and $A$ aren't preparable simultaneously in sharp states. Moreover, with the additional conditions for decision-supporting superinformation theories (see section \ref{decision}), the states $a_{\pm}$ represent non-sharp states in an $X$-indistinguishability class with equal weights for $x_0$ and $x_1$.	Thus we can define the new task $\mathcal{W}$ as
	\begin{equation}\label{W}
	\mathcal{W}=\{x_0 \rightarrow a_{+}, x_1 \rightarrow a_{-}\}
	\end{equation}
	This task, which is a special kind of distinguishing task,  can be considered as the constructor-theoretic generalization of the Walsh-Hadamard gate in quantum computation context. The transpose (the task with all the input and output states swapped) of $\mathcal{W}$ is of course $\mathcal{W}^{\sim}=\{a_{+} \rightarrow x_0, a_{-} \rightarrow x_1  \}$. What we are describing is a reversible physical transformation which transforms some sharp states into non-sharp states, and viceversa.  We underline the fact that the coin task $\mathcal{C}$, differently from $\mathcal{W}$, $\mathcal{W}^{\sim}$, doesn't admit the transpose. 
\end{itemize}

The existence of $\mathcal{W}$ task is very important in studying emotions.  It describes a typical evolution of subject's belief about an observable from centainty to uncertainty (and viceversa): for example if we toss a coin which has an initial value $x_0$ (or $x_1$), subject's belief about the outcomes will be a generalized mixture of $x_0$ and $x_1$ with two different affective values.
	
However, it is clear that there may be other possible evolutions of the initial state. In the following section, we will introduce another important element to better understand emotions.

\subsection{The phase task and emotion valence}\label{task_phase}
We now introduce in constructor theoretic form a task which is very important in quantum computation, the \textit{phase task}: given a substrate $S$, an observable $X$ and a state with a specific $X$-partition of unity, the phase task $\mathcal{F}$  acts on such state by leaving unchanged the original partition of unity. In other words, the phase task defines a class of trasformations of states which leaves them in the same $X$-indistinguishability class. 
\begin{equation}\label{phase_task}
\mathcal{F}_X: \cup_{x \in X} \{s \rightarrow s': [f_x]_{s} =  [f_x]_{s'}\}
\end{equation}
Let us now examine some examples:
\begin{itemize}
	\item Clearly the identity task is a trivial example of the phase task.
	\item 	$\mathcal{F}_X$ applied to a sharp state of $X$ (for example $x_0$) can change the state only in a trivial way, since sharp states also exist in classical information.
	\item Given the observables $\{x_0, x_1\}$ and $\{a_+, a_-\}$ previously defined, there exists a specific phase task which produces the following transformation $\mathcal{F}_X(a_+)=a_-$, thus swapping states $a_{\pm}$. 
\end{itemize}

In other words, the phase task does not change the form of a generalized mixture in terms of $X$-partition, and thus it apparently does not change judgements. However, it introduces an affective change, which will manifest when we apply subsequently different tasks. We thus consider some interesting serial composition of tasks $\mathcal{W}^{\sim} \mathcal{F}\mathcal{W}$ and we show how they are able to  change the $X$-partition of unity.
\begin{itemize}
	\item  When $\mathcal{F}=I$, we simply have $\mathcal{W}^{\sim} W= I$: the serial application of task $W$ and its transpose is clearly the identity task. This means that subjects can mentally simulate a coin toss and then revert the generalized mixture into the original sharp state, without any affective change;
	\item When $\mathcal{F}$ is a swap of states $y_{\pm}$, the sharp  input states  $\{x_0,x_1\}$ are swapped. This can be shown by considering for example an input state $x_0$, which is transfomed first in $a_+$, then swappend into $a_-$ and finally into $x_1$. In this case the $X$-partition of unity changes from parameter $\theta =0$ (for example for input sharp state $x_0$) to $\theta = \pi$ (output state $x_1$). Thus the phase task has applied a phase shift of $\pi$. This represents a complete belief change about $X$ due to an emotional change;
	\item We can consider a general  case where $\mathcal{W}^{\sim} \mathcal{F}\mathcal{W}$ changes the initial state with $X$-partition of unity 
	$[f_{x_0},f_{x_1}]=[cos^2(\theta/2), sin^2(\theta/2)]$ into another state with partition $[cos^2(\theta'/2), sin^2(\theta'/2)]$.
\end{itemize}
We can thus conclude that the phase task can be described in terms of a parameter, the phase $\theta$, which acts, when combined with $\mathcal{W}$ task, by changing the $X$-partition of unit. 
The phase task allows us to describe emotions in constructor theoretical terms: a change of emotion valence  is put in correspondence with a specific phase task $\mathcal{F}$, which, when combined with a second observable through task $\mathcal{W}$, leads to a change of judgements of such observable. 

\section{Low affect infusion strategies}
In low affect infusion situations, subjects can manage feelings and belief in an independent way. Moreover, to perform judgements about an event, they use only information relevant to that event. 

Let  $J(x)$ be  the subject's judged likelihood about event $x$.  Affect valence $a$ is a variable really \textit{independent} from the external event $x$: in terms of judgements, this means that $J(x)=J(x|a)$. In constructor theoretic terms, we can say that the observables relevant to the event $X$ and to the affect valence $A$ form a classic information medium. 

Rational subjects  perform judgements only based on similar pre-stored judgements (for example $J(x) \simeq J(k)$ for $K$ and $X$  observables representing similar situations), or on specific information about the object of judgment (motivated processing). Thus the observable core affect is not used to perform judgements in this case, since it  has no bearing in the judged object.

\subsection{Classic information and emotions}
Constructor theory of information helps us to analyze the low affect infusion strategy in terms of observables and their relations. Observables $X$ and $A$ form a classic substrate. According to this strategy, subjects are able in principle to perform  judgements about their affective state $A$ or about the event $X$ without mixing them.

A typical technique used to disentangle affect and knowledge is based on expliciting \textit{how do I feel}: in this way,  subjects are helped to manage affect as an independent observable.
It is clear that this strategy requires the use of classical information. According to constructor theory of information, observables $A$ (affect valece) and $X$ (an event) can be managed together and also their combination is an observable. We can clearly consider conditional tasks, as in section (\ref{task_conditional}): the conditionsal task $\mathcal{CT}(X,A)$ defines a situation where the specific knowledge of $X$ modifies the affect: for example, a bad news can change the affect valence, but in any moment the subject is able to perform measurements about observables $X$ and $A$. In this sense, this is a \textit{rational-like} situation, because affect, even present, can be managed separately. 
	
We can clearly apply the same general considerations used in section (\ref{inf_superinf}), where we defined a simple formula (\ref{conj_general}) which is required for judgements in rational regime. Thus we have for judgements relevant to $A$ and $X$ forming classic information media the condition: $J(x,a)\leq J(x)$ and $J(x,a) \leq J(a)$. For example, subjects are able to judge more likely to win a lottery than winning that lottery and at the same time being happy.  Clearly such judgements require to use rational strategies, and are incompatble with high affect infusion strategires, as we will see later.

\section{High affect infusion strategies}

\subsection{Heuristic processing}
We first recall from \cite{forgas2002managing} the first high affect infusion strategy, the \textit{heuristic processing}: it refers to constructive but truncated, low-effort processing, which is likely to be adopted when time and personal resources such as motivation, interest, attention, and working-memory capacity are scarce (e.g., evaluating your friend’s new company car). Heuristic processing may result in mood congruence when affect
is used as a heuristic cue and it is consistent   with  the \textit{‘affect-as-information}’ hypothesis \cite{emotclore2007}. According to such hypothesis, affect provides compelling information about the personal value of whatever is in mind at the time. In other words, when making evaluative judgments, people often ask
themselves ‘\textit{how do I feel about it}?’. In such cases, positive affect signals that the object of judgment is valuable, leading to a positive evaluation, and negative affect signals that it lacks value, leading to a negative evaluation.  Thus, feelings may be used as a shortcut to produce a judgement on a given target, as long as the feeling is perceived (rightly or wrongly) to be evoked by the target object.

We now analyze such heuristic processing in constructor theoretic terms. We first note that the  \textit{‘affect-as-information}’ hypothes can be formally written in the form of the following equation
\begin{equation}\label{symmetry}
	J(x|a) \simeq J(a|x)
\end{equation}
 In other words, the affect judgement when assuming an event can be used as a judgement of the same event when assuming that affect.

This strategy has been found in other contexts, like for example in the \textit{inverse fallacy} \cite{meehl1955antecedent, hammerton1973case, liu1975specific, eddy1982probabilistic} for which a simple quantum-like model has been proposed \cite{franco2007inverse}: the symmetry of the inner product and a specific structure of the vector space enconding information is able to reproduce this effect.

While in low affecty infusion strategies we can simply refer to classical information medium, we now show that the strategy defined by equation (\ref{symmetry}) requires the use of superiformation. Let us suppose that $X$ and $A$ form an information medium and that conditions for decision-supporting superinformation theories are true: then $J(x,a)=J(x|a)J(a)=J(a|x)J(x)$. Thus the symmetry property (\ref{symmetry}) is true only when $J(x)=J(a)$, which is without meaning, because affect $A$ and the event $X$ mustbe independent.

\subsection{Other high influence processes - emotions and memory}
High affect infusion strategies are more complex than simple heuristic processing: for example  \textit{affect priming}, which suggests that affect increases access to congruent memories, is a kind of high influence process. As evidenced in \cite{seo2007being}, affective feelings can hurt or help decision making: this fact can be largely determined by how individuals experience and handle those feelings in more or less functional or dysfunctional ways: according to such \textit{feeling-as-decision-facilitator} perspective, affective feelings can improve decision-making performance by facilitating and even enabling decision-making processes. 

In order to describe such mechanisms, we have to introduce in a general way the relation between affect and knowledge.
Let be $X$ and $A$  \textit{complementary observables} of a superinformation medium. Clearly, we thus exclude situations where both affect and the object are sharp states in subject's brain substrates. On the contrary, we consider situations where $X$ and $A$ together can't form a clonable observable: a subject can't know them both simultaneously, and a measurement of (for example) $A$  changes the context and thus the real condition of $X$. This condition is clearly related to the concept of emotional truth \cite{de2002emotional}: it is well known that there’s a big difference between emotional and factual truth of an experience or situation. In particular, we identified the task  $\mathcal{W}^{\sim} \mathcal{F}\mathcal{W}$ described in section (\ref{task_phase}) as the task leading to emotional truth, driven by emotion phase parameter in $\mathcal{F}$.

Two previous articles
\cite{franco2008grover, franco2009availability} attempted to use quantum search algorithms in the context of human memory. The first considered  phase matching conditions known for generalized Grover's algorithm, connecting such phase to emotion regulation strategies. The second article used another kind of generalization, the amplitide amplification algorithm, to explain a cognitive heuristic called availability bias. In particular,  the ease of
a memory task and the estimated probabilty were connected together, with a quadratic relation clearly connected with the quantum speedup.
In this section, we focus the attention of the internal mechanism of such algorithms, since we now have  a solid constructor theoretic definion of phase and emotion. At the core of such generalized algorithms \cite{long2002phase}, we can always find the repeated use of an operator $-I_{\gamma}(\theta) U^{-1}I_{\tau}(\phi)U$, where $-I_{\gamma}(\theta)$ and $I_{\tau}(\phi)$ are inversion operators for specific states and with specific angles. We recognize in $U^{-1}I_{\tau}(\phi)U$ the quantum-computational version of the serial task composition $\mathcal{W}^{\sim} \mathcal{F}\mathcal{W}$ described in section (\ref{task_phase}). The main result in \cite{long2002phase} is that phase matching between the phase rotation of the marked
state and the prepared state $\gamma$ is crucial in constructing a quantum search algorithm. 

Thus we conclude that the emotional nature of phase rotation task allows us to explain the mood-congruent mechanism (for a review, see \cite{clore2007emotions}) by using the constructor-theoretic generalizations of  Grover's algorithm  together with the emotional interpretation of phase task.

\section{Conclusions}
Even if there is a prior attempt to use quantum concepts to describe emotions \cite{lukac2007quantum, yan2015bloch, yan2019quantum}
(more specifically, a model of robot-specific emotions based on quantum cellular automata and Bloch sphere-based representations), the approach proposed in the present article is more general, since it is well grounded in constructor theory of information. 

We recall here the main results of this article: 1) the combination of constructor theoretic concepts with the theory of constructed emotion. In particular, the identification of core affect valence as a first  observable to study - from a constructor theoretic approach - as a physical variable;
2) the demonstraton that there are specific situations where core affect valence and the knowledge of events require the use of superinformation;
3) the use of additional hypotheses, taken form the decision supporting argument, to describe emotions as a complementary observable, and the definition of a phase task;
4) the connection between the phase task and emotion, thus  recognizing the role of emotions is quantum Grover's algorithm and its generalizations.

In other words, this article shows, under precise and general assumptions, what has been hypotesized in previous articles \cite{franco2008grover, franco2009availability}: emotions cannot be  studied and understood without using new conceptual instruments, like those introduced by superinformation. Moreover, this approach shows how emotions and cognitive tasks like memory and judgement are tightly correlated.

\footnotesize
%
%

\section{References}
\bibliography{../mybib}{}

\begin{thebibliography}{10}

\bibitem{deutsch2013constructor}
David Deutsch.
\newblock Constructor theory.
\newblock {\em Synthese}, 190(18):4331--4359, 2013.

\bibitem{barrett2007mice}
Lisa~Feldman Barrett, Kristen~A Lindquist, Eliza Bliss-Moreau, Seth Duncan,
  Maria Gendron, Jennifer Mize, and Lauren Brennan.
\newblock Of mice and men: Natural kinds of emotions in the mammalian brain? a
  response to panksepp and izard.
\newblock {\em Perspectives on Psychological Science}, 2(3):297--312, 2007.

\bibitem{franco2019first}
Riccardo Franco.
\newblock First steps to a constructor theory of cognition.
\newblock {\em arXiv preprint arXiv:1904.09829}, 2019.

\bibitem{busemeyer2011quantum}
Jerome~R Busemeyer, Emmanuel~M Pothos, Riccardo Franco, and Jennifer~S
  Trueblood.
\newblock A quantum theoretical explanation for probability judgment errors.
\newblock {\em Psychological review}, 118(2):193, 2011.

\bibitem{franco2016newtheory}
Riccardo Franco.
\newblock Towards a new quantum cognition model.
\newblock {\em arXiv preprint arXiv:1611.09212v1}, 2016.

\bibitem{james1894discussion}
William James.
\newblock Discussion: The physical basis of emotion.
\newblock {\em Psychological review}, 1(5):516, 1894.

\bibitem{lange1922emotions}
Carl~Georg Lange and Istar~A Haupt.
\newblock The emotions.
\newblock 1922.

\bibitem{cannon1927james}
Walter~B Cannon.
\newblock The james-lange theory of emotions: A critical examination and an
  alternative theory.
\newblock {\em The American journal of psychology}, 39(1/4):106--124, 1927.

\bibitem{schachter1962cognitive}
Stanley Schachter and Jerome Singer.
\newblock Cognitive, social, and physiological determinants of emotional state.
\newblock {\em Psychological review}, 69(5):379, 1962.

\bibitem{arnold1960emotion}
Magda~B Arnold.
\newblock Emotion and personality.
\newblock 1960.

\bibitem{emotlazarus1991emotion}
Richard~S Lazarus.
\newblock {\em Emotion and adaptation}.
\newblock Oxford University Press on Demand, 1991.

\bibitem{ekman1999basic}
Paul Ekman.
\newblock Basic emotions.
\newblock {\em Handbook of cognition and emotion}, 98(45-60):16, 1999.

\bibitem{barrett2006emotions}
Lisa~Feldman Barrett.
\newblock Are emotions natural kinds?
\newblock {\em Perspectives on psychological science}, 1(1):28--58, 2006.

\bibitem{barrett2006solving}
Lisa~Feldman Barrett.
\newblock Solving the emotion paradox: Categorization and the experience of
  emotion.
\newblock {\em Personality and social psychology review}, 10(1):20--46, 2006.

\bibitem{clore2007emotions}
Gerald~L Clore and Jeffrey~R Huntsinger.
\newblock How emotions inform judgment and regulate thought.
\newblock {\em Trends in cognitive sciences}, 11(9):393--399, 2007.

\bibitem{cacioppo2000psychophysiology}
John~T Cacioppo, Gary~G Berntson, Jeff~T Larsen, Kirsten~M Poehlmann, Tiffany~A
  Ito, et~al.
\newblock The psychophysiology of emotion.
\newblock {\em Handbook of emotions}, 2:173--191, 2000.

\bibitem{emotcasper2001affective}
Karen Casper.
\newblock Affective feelings as feedback: Some cognitive consequences.
\newblock {\em Theories of mood and cognition: A user’s guidebook}, 27, 2001.

\bibitem{russell2003core}
James~A Russell.
\newblock Core affect and the psychological construction of emotion.
\newblock {\em Psychological review}, 110(1):145, 2003.

\bibitem{russell1999core}
James~A Russell and Lisa~Feldman Barrett.
\newblock Core affect, prototypical emotional episodes, and other things called
  emotion: dissecting the elephant.
\newblock {\em Journal of personality and social psychology}, 76(5):805, 1999.

\bibitem{blanchette2010influence}
Isabelle Blanchette and Anne Richards.
\newblock The influence of affect on higher level cognition: A review of
  research on interpretation, judgement, decision making and reasoning.
\newblock {\em Cognition \& Emotion}, 24(4):561--595, 2010.

\bibitem{meiselman2016emotion}
Herbert~L Meiselman.
\newblock {\em Emotion measurement}.
\newblock Woodhead publishing, 2016.

\bibitem{lindquist2012brain}
Kristen~A Lindquist, Tor~D Wager, Hedy Kober, Eliza Bliss-Moreau, and
  Lisa~Feldman Barrett.
\newblock The brain basis of emotion: a meta-analytic review.
\newblock {\em The Behavioral and brain sciences}, 35(3):121, 2012.

\bibitem{emotclore2007}
Gerald~L Clore and Jeffrey~R Huntsinger.
\newblock How emotions inform judgment and regulate thought.
\newblock {\em Trends in cognitive sciences}, 11(9):393--399, 2007.

\bibitem{johnson1983affect}
Eric~J Johnson and Amos Tversky.
\newblock Affect, generalization, and the perception of risk.
\newblock {\em Journal of personality and social psychology}, 45(1):20, 1983.

\bibitem{kadous2001improving}
Kathryn Kadous.
\newblock Improving jurors' evaluations of auditors in negligence cases.
\newblock {\em Contemporary Accounting Research}, 18(3):425--444, 2001.

\bibitem{schwarz1983mood}
Norbert Schwarz and Gerald~L Clore.
\newblock Mood, misattribution, and judgments of well-being: informative and
  directive functions of affective states.
\newblock {\em Journal of personality and social psychology}, 45(3):513, 1983.

\bibitem{gorn2001arousal}
Gerald Gorn, Michel Tuan~Pham, and Leo Yatming~Sin.
\newblock When arousal influences ad evaluation and valence does not (and vice
  versa).
\newblock {\em Journal of consumer Psychology}, 11(1):43--55, 2001.

\bibitem{muris2006anxiety}
Peter Muris and Simone van~der Heiden.
\newblock Anxiety, depression, and judgments about the probability of future
  negative and positive events in children.
\newblock {\em Journal of anxiety disorders}, 20(2):252--261, 2006.

\bibitem{forgas2002managing}
Joseph~P Forgas and Joseph~V Ciarrochi.
\newblock On managing moods: Evidence for the role of homeostatic cognitive
  strategies in affect regulation.
\newblock {\em Personality and Social Psychology Bulletin}, 28(3):336--345,
  2002.

\bibitem{forgas1995mood}
Joseph~P Forgas.
\newblock Mood and judgment: the affect infusion model (aim).
\newblock {\em Psychological bulletin}, 117(1):39, 1995.

\bibitem{segal1987ellsberg}
Uzi Segal.
\newblock The ellsberg paradox and risk aversion: An anticipated utility
  approach.
\newblock {\em International Economic Review}, pages 175--202, 1987.

\bibitem{sarin1998revealed}
Rakesh Sarin and Peter Wakker.
\newblock Revealed likelihood and knightian uncertainty.
\newblock {\em Journal of Risk and Uncertainty}, 16(3):223--250, 1998.

\bibitem{marletto2016constructor}
Chiara Marletto.
\newblock Constructor theory of probability.
\newblock 472(2192):20150883, 2016.

\bibitem{smith1984conceptual}
Edward~E Smith and Daniel~N Osherson.
\newblock Conceptual combination with prototype concepts.
\newblock {\em Cognitive science}, 8(4):337--361, 1984.

\bibitem{festinger1962theory}
Leon Festinger.
\newblock {\em A theory of cognitive dissonance}, volume~2.
\newblock Stanford university press, 1962.

\bibitem{meehl1955antecedent}
Paul~E Meehl and Albert Rosen.
\newblock Antecedent probability and the efficiency of psychometric signs,
  patterns, or cutting scores.
\newblock {\em Psychological bulletin}, 52(3):194, 1955.

\bibitem{hammerton1973case}
M~Hammerton.
\newblock A case of radical probability estimation.
\newblock {\em Journal of Experimental Psychology}, 101:252, 1973.

\bibitem{liu1975specific}
An-Yen Liu.
\newblock Specific information effect in probability estimation.
\newblock {\em Perceptual and Motor Skills}, 41(2):475--478, 1975.

\bibitem{eddy1982probabilistic}
David~M Eddy.
\newblock Probabilistic reasoning in clinical medicine: Problems and
  opportunities.
\newblock 1982.

\bibitem{franco2007inverse}
Riccardo Franco.
\newblock The inverse fallacy and quantum formalism.
\newblock {\em arXiv preprint arXiv:0708.2972}, 2007.

\bibitem{seo2007being}
Myeong-Gu Seo and Lisa~Feldman Barrett.
\newblock Being emotional during decision making—good or bad? an empirical
  investigation.
\newblock {\em Academy of Management Journal}, 50(4):923--940, 2007.

\bibitem{de2002emotional}
Ronald De~Sousa.
\newblock Emotional truth: Ronald de sousa.
\newblock In {\em Aristotelian Society Supplementary Volume}, volume~76, pages
  247--263. Wiley Online Library, 2002.

\bibitem{franco2008grover}
Riccardo Franco.
\newblock Grover's algorithm and human memory.
\newblock {\em arXiv preprint arXiv:0804.3294}, 2008.

\bibitem{franco2009availability}
Riccardo Franco.
\newblock Quantum amplitude amplification algorithm: an explanation of
  availability bias.
\newblock In {\em International Symposium on Quantum Interaction}, pages
  84--96. Springer, 2009.

\bibitem{long2002phase}
Gui-Lu Long, Xiao Li, and Yang Sun.
\newblock Phase matching condition for quantum search with a generalized
  initial state.
\newblock {\em Physics Letters A}, 294(3-4):143--152, 2002.

\bibitem{lukac2007quantum}
Martin Lukac and Marek Perkowski.
\newblock Quantum mechanical model of emotional robot behaviors.
\newblock In {\em 37th International Symposium on Multiple-Valued Logic
  (ISMVL'07)}, pages 19--19. IEEE, 2007.

\bibitem{yan2015bloch}
Fei Yan, Abdullah~M Iliyasu, Zhen-Tao Liu, Ahmed~S Salama, Fangyan Dong, and
  Kaoru Hirota.
\newblock Bloch sphere-based representation for quantum emotion space.
\newblock {\em Journal of Advanced Computational Intelligence and Intelligent
  Informatics}, 19(1):134--142, 2015.

\bibitem{yan2019quantum}
Fei Yan, Abdullah~M Iliyasu, Sihao Jiao, and Huamin Yang.
\newblock Quantum structure for modelling emotion space of robots.
\newblock {\em Applied Sciences}, 9(16):3351, 2019.

\end{thebibliography}
\bibliographystyle{unsrt}
\end{document}